\documentclass[conference]{IEEEtran}
\IEEEoverridecommandlockouts
\usepackage{cite}
\usepackage{amsmath,amssymb,amsfonts}
\usepackage{algorithmic}
\usepackage[ruled,vlined,linesnumbered]{algorithm2e}
\usepackage{graphicx}
\usepackage{textcomp}
\usepackage{xcolor}
\usepackage{url}
\usepackage{footnote}
\usepackage[perpage]{footmisc}
\usepackage{balance}
\usepackage{boldline, multirow}
\usepackage{listings}
\usepackage{tabularx,booktabs,ragged2e,float,adjustbox,makecell,array}
\usepackage{csquotes}
\usepackage{hyperref}
\usepackage{tikz}
\usepackage{pifont}
\newcommand{\xmark}{\ding{55}}
\usepackage{comment}
\usepackage[table]{colortbl}
\usepackage{hhline}
\usepackage{tcolorbox}
\usepackage[normalem]{ulem}
\usepackage[normalem]{ulem}
\def\BibTeX{{\rm B\kern-.05em{\sc i\kern-.025em b}\kern-.08em
    T\kern-.1667em\lower.7ex\hbox{E}\kern-.125emX}}
    
\begin{document}

\newcommand{\SkipBeforeAndAfter}{\vspace{-1em}}
\newcommand{\tool}[1]{\textit{RLMutation}}

\newcommand{\Foutse}[1]{\textcolor{red}{{\it[Foutse:#1]}}}
\newcommand{\Amin}[1]{\textcolor{blue}{{\it[Amin:#1]}}}
\newcommand{\Florian}[1]{\textcolor{orange}{{\it[Florian:#1]}}}
\newcommand{\Ahura}[1]{\textcolor{teal}{{\it[Ahura:#1]}}}
\newcommand{\AhuraTODO}[1]{\textcolor{olive}{{\it[Ahura TODO:#1]}}}

\newcommand{\eg}{\textit{e}.\textit{g}., }
\newcommand{\ie}{\textit{i}.\textit{e}., }

\newcommand{\rqone}{What are the limitations of existing mutation killing definitions when applied to RL?}
\newcommand{\rqtwo}{How are different agents and environments affected by the different mutations?}
\newcommand{\rqthree}{Do the HOM generated from our FOM possess the subsuming property similarly to traditional software engineering?}

\title{Mutation Testing of Deep Reinforcement Learning Based on Real Faults\\
}

\author{\IEEEauthorblockN{Florian Tambon\textsuperscript{\textsection},Vahid Majdinasab\textsuperscript{\textsection}, Amin Nikanjam, Foutse Khomh, Giuliano Antoniol}
\IEEEauthorblockA{\textit{Polytechnique Montréal, Canada} \\
\{florian-2.tambon, vahid.majdinasab, amin.nikanjam, foutse.khomh, giuliano.antoniol\}@polymtl.ca}
}
\maketitle
\thispagestyle{plain}
\pagestyle{plain}
\begingroup\renewcommand\thefootnote{\textsection}
\footnotetext{Equal contribution}

\begingroup\renewcommand\thefootnote{\textsection}

\begin{abstract}
Testing Deep Learning (DL) systems is a complex task as they do not behave like traditional systems would, notably because of their stochastic nature. Nonetheless, being able to adapt existing testing techniques such as Mutation Testing (MT) to DL settings would greatly improve their potential verifiability. While some efforts have been made to extend MT to the Supervised Learning paradigm, little work has gone into extending it to Reinforcement Learning (RL) which is also an important component of the DL ecosystem but behaves very differently from SL. This paper builds on the existing approach of MT in order to propose a framework, \tool{}, for MT applied to RL. Notably, we use existing taxonomies of faults to build a set of mutation operators relevant to RL and use a simple heuristic to generate test cases for RL. This allows us to compare different mutation killing definitions based on existing approaches, as well as to analyze the behavior of the obtained mutation operators and their potential combinations called Higher Order Mutation(s) (HOM). We show that the design choice of the mutation killing definition can affect whether or not a mutation is killed as well as the generated test cases. Moreover, we found that even with a relatively small number of test cases and operators we manage to generate HOM with interesting properties which can enhance testing capability in RL systems.
\end{abstract}


\section{Introduction}\label{sec:intro}
Mutation Testing is a white box testing method that aims to inject \textit{artificial} changes based on \textit{real} faults in order to evaluate a test suite's capability to reveal faults. Mutation Testing has been extensively studied and used in traditional software engineering \cite{Jia11, Papadakis19} to assess the quality of test suites. A fundamental hypothesis of Mutation Testing is the Coupling Effect hypothesis which posits that \enquote{complex mutants are coupled to simple mutants in such a way that a set of test cases that detects all simple mutants in a program will detect a \textbf{large percentage} of the complex mutants} \cite{Offutt92}. As such uncovering such unkilled complex mutants is of interest, with one way of generating such complex mutants being to combine simple mutants, called First Order Mutation(s) (FOM), together. This is the concept of Higher Order Mutation(s) (HOM) introduced by Jia et al. \cite{Jia09}.

Such established testing techniques and concepts could be useful for Deep Learning systems, in an effort to increase their reliability, since such systems are notoriously hard to test because of their peculiar nature \cite{Zhang22}. Because of the paradigm differences between Deep Learning-based systems and traditional software systems, it is only recently that researchers have started proposing Mutation Testing frameworks tailored for Deep Learning-based systems, in particular Supervised Learning \cite{Ma18, Hu19, humbatova2021deepcrime}, to assess the quality of test dataset at revealing faults.

Yet, Supervised Learning is not the only sub-paradigm in Machine Learning, and (deep) Reinforcement Learning (RL), one of the other main sub-paradigms with a wide range of applications \cite{Luong19, Polydoros17} is increasingly being adopted in practice. RL differs deeply from Supervised Learning: while in Supervised Learning a model learns from a training dataset in order to generalize to any new data from the input distribution, RL is based on the idea of training an agent using its interaction with an environment through a feedback system \cite{LavetDRL2018}. For instance, a robot (agent) evolving in a room (environment) with a goal to go from A to B with some traps on the way. As such, previously introduced frameworks might present several limitations if applied to RL, for instance, the mutation operators defined in Supervised Learning to obtain mutant models might not apply to RL. In parallel, to the best of our knowledge, \cite{Lu2022-RLmutation} is the only research work that tackled Mutation Testing in RL; proposing a fault detection approach that is based on the manual crafting of relevant environments. In particular, they introduced the idea that traditional test cases used in Mutation Testing applied to traditional software systems could be translated to the notion of test environments in RL. However, their study is limited to only one type of RL algorithm and does not explore real fault-based operators nor the potential usefulness of combining existing operators to form HOM which could prove useful to assess test environments' capacity to find subtle faults in RL systems.

In this paper, we propose a framework, \tool~, for Mutation Testing of deep RL programs leveraging HOM adapted to RL. We defined mutation operators for RL motivated by existing taxonomized faults. We then analyzed how they fared on different RL environments and algorithms by using and comparing a number of mutation killing definitions adapted from previous works. In order to leverage HOM power to highlight more complex faults, we adapt existing work on HOM \cite{Jia09} to the RL task specificity. Namely, we conceive a simple heuristic tailored to the RL problem to systematically generate some test environments in order to obtain test cases to assess HOM usefulness. Thus, we aim to provide some insights into how Mutation Testing could be applied to RL while acknowledging existing differences with Supervised Learning.

Our contribution is the proposed RL framework composed of the following: $11$ mutation operators based on real taxonomized faults, a comparison of the impact of mutation killing definition design over the FOM killed, and a heuristic to generate relevant test environments to study both FOM and HOM.
 
\textbf{The remainder of the paper is structured as follows:} Section \ref{sec:background} gives relevant background knowledge about Mutation Testing, HOM, and RL. Section \ref{sec:mutation_op} presents the mutation operators introduced, as well as the procedure to determine how to generate relevant HOM when using Mutation Testing for RL. Section \ref{sec:res} reports about our experiments and results. Section \ref{sec:threats} discusses threats to the validity of our work. Section \ref{sec:rel} reviews the related literature, while Section \ref{sec:concl} concludes the paper.

\section{Background}\label{sec:background}
\subsection{Reinforcement Learning} \label{subsec:rl_backgorund}
RL is a Machine Learning sub-paradigm in which an \textit{agent} learns from interacting with an \textit{environment} \cite{LavetDRL2018}. The main task consists of learning how to map states to actions by maximizing the long-term reward. So, there are 3 main components in an RL problem: environment, agent, and a learned policy. Formally, at each time step $t$, the agent perceives the state of the environment it is in ($s_t \in S$). After doing so, the agent takes an action ($a_t \in A$). Upon taking the action, the agent receives a reward ($r_t \in R$), and transitions to the next state ($s_{t+1} \in S$). The symbols $S$, $A$, and $R$ denote the state, action, and reward spaces, respectively.

We denote the $(s_t, a_t, r_t, s_{t+1})$ tuple, which contains a record of the agent's interaction with the environment as an \textit{observation}. The agent interacts with the environment and collects observations until the environment reaches a \textit{terminal} state where the environment ends. An \textit{episode} indicates the length of the experiences the agent collects starting from an (initial) state and ending in a terminal state. The agent aims to learn a policy $\pi \in \Pi$ (the policy space) which maximizes the expected cumulative reward or return. 

As the agent interacts with the environment and collects rewards, it needs to learn the trade-off between short-term and long-term rewards. To facilitate this concept, the \textit{discount factor} ($\gamma$), a hyperparameter between 0 and 1, is used during the calculation of the cumulative rewards \cite{sutton2018reinforcement}. Equation \ref{eqn:policy_value_formula} shows how the value of a policy in a state is calculated:
\begin{equation}\label{eqn:policy_value_formula}
    V^\pi(s) = \mathbb{E}[\,R_t | s_t = s\,], \text{with}\,R_t = \sum_{k=0}^{\infty} \gamma^k r_{t+k+1}
\end{equation}

Recently, researchers have successfully integrated Deep Learning methods in RL to solve some challenging sequential decision-making problems \cite{Goodfellow-et-al-2016} by employing Deep Neural Networks to learn the policy effectively \cite{mnih2015human,gandhi2017learning, moravvcik2017deepstack}.


\subsection{Mutation Testing and Higher Order Mutation}\label{sec:background:hom}

Jia et al. \cite{Jia09} studied the different types of HOM, by classifying them based on two main properties: 1) a HOM is \textit{Coupled} if, for a given set of test cases killing the HOM $T_H \neq \emptyset$ and a given union of sets of test cases killing its constituent FOM $\underset{i}{\cup} ~ T_i$, we have $T_H \cap \underset{i}{\cup} ~ T_i \neq \emptyset$, and 2) HOM can be \textit{Subsuming} if $|T_H| < |\underset{i}{\cup} ~ T_i|$ where $|~.~|$ is the cardinal of the set. The latter can be further refined between Strongly and Weakly Subsuming, with Strongly Subsuming HOM being defined as $T_H \subset \underset{i}{\cap} ~ T_i$. From the definitions, one can see that Subsuming HOM are of particular interest as they lead to mutations that turn out to be more complex to be detected than their constituents FOM. Thus, focusing on HOM means fewer and more subtle mutants to use in Mutation Testing.

\subsection{Mutation Testing of Deep Learning}\label{sec:background:Mutation Testing_ml}

Applying Mutation Testing to Deep Learning, similarly to how it is done in traditional software engineering, raises several questions with the most prominent one being: Is a test case killing a mutation, or is it just an artifact of the stochasticity of the model's training? Indeed, given a test case $x$, a neural network $N$ and a mutant $M$, having $N(x) \neq M(x)$ does not necessarily mean that $x$ killed the mutation $M$, as the mutation being killed could only be due to the stochasticity inherent to the training of the neural network. In particular, training two neural networks, $N_1$ and $N_2$, on the same data and specification does not guarantee they will agree on a test case, as multiple sources of stochasticity might make them diverge from each other. Thus, previous researchers in Supervised Learning made an argument in favor of considering not just one instance of a model but rather a group of instances when using Mutation Testing \cite{Ma18, humbatova2021deepcrime}.

DeepMutation \cite{Ma18} introduced the idea of averaging the kill ratio of multiple versions of a model (i.e., neural network). Jahangirova et al. \cite{Jahangirova20} and then Humbatova et al.  \cite{humbatova2021deepcrime} proposed to consider Mutation Testing through the lens of statistical testing over the accuracy of a distribution of instances that compares $n$ non-mutant instances against $n$ mutant instances. In particular, for a given test set $\mathcal{T}$ and the sets of accuracy over $\mathcal{T}$ of non-mutated models ($A_{N_1}, ..., A_{N_n}$) and mutated models with a given mutation operator ($A_{M_1}, ..., A_{M_n}$), they defined the following test function:

\begin{equation}
MT_\mathcal{T} = \begin{cases}
1 & \text{if p-value} < 0.05 ~ \text{and effectSize} \geq 0.5\\
0 & \text{else}
\end{cases}
\end{equation}
where the \textit{p-value} is obtained by using \textit{Generalised Linear Model} \cite{Nelder72} and the \textit{effectSize} is calculated using \textit{Cohen's d} \cite{Kelley12}. While not directly part of the test function, they consider the power analysis to exclude mutations for which the statistical power of the test is too low (with the threshold $\beta \geq 0.8$).

Mutation Testing has also been applied to RL. Authors in \cite{Lu2022-RLmutation} proposed an approach to evaluate a crafted environment's ability for revealing mutants based on mutation operators designed for RL. They defined a mutant as killed if, for a healthy agent $A$ and a mutated agent $A_M$, the ratio $p_M/p$ of their average rewards over $n$ episodes on a given environment $E$ are inferior to a given threshold $\theta$. As such, one can see that just like with Supervised Learning, there can be cases where $(A_1$,$A_{M_1})$ would reveal the mutation, yet $(A_2$,$A_{M_2})$ would not because of RL's inherent stochasticity, which can deeply change the results among trained agents using the same environment and hyperparameter configurations but with different seeds \cite{Henderson18, Agarwal21}. Hence, two users training two agents on the same specification would end up with two different test results.

\section{Mutation operator for (Deep) Reinforcement Learning}\label{sec:mutation_op}

\subsection{First Order Mutation}

Lu et al. introduced several mutation operators applicable to RL \cite{Lu2022-RLmutation}. Nonetheless, their operators present several shortcomings. First, some can not be generalized to any RL algorithm which limits their relevance, for instance, mutations based on the \textit{epsilon} parameter are limited to a subset \textit{Off-policy} algorithms such as \textit{DQN}. Secondly, some operators such as removing/adding a neuron on a particular layer will most likely lead to a crash, as the fault itself leads to cascading changes in the model architecture as pointed out in DeepCrime \cite{humbatova2021deepcrime}. Finally, some of the mutation operators are not justified since they are not based on real faults which undermine the usefulness of the operator, for instance, \textit{shuffling the replay priority} in the replay buffer is not motivated by any real fault.

Thus, to obtain relevant FOMs, we started from operators defined in \cite{Lu2022-RLmutation} as a basis and further extracted existing taxonomies reporting on bugs affecting RL or Deep Learning models \cite{nikanjam2022faults}\cite{humbatova2020taxonomy} or directly adapting existing mutation operators \cite{Jahangirova20}\cite{humbatova2021deepcrime} used in Supervised Learning. We obtained a (non-exhaustive) list of mutation operators that we divided into three categories (i.e., environment, agent, and policy) based on what the mutation is affecting. Due to space constraints, we only briefly describe the mutation operators in this section. A comprehensive description can be found in our replication package \cite{rep-package} with a reference to the specifics that motivated each operator along with a comparison with operators defined in \cite{Lu2022-RLmutation}.

\textbf{Environment-level:} The environment-level mutations are meant to simulate the faults that can happen when an agent receives observations as it interacts with the environment, i.e. receiving an incorrect observation. These faults can, for instance, be the results of faulty sensors, faults in environment design, or even nefarious attacks \cite{Everitt17, Romoff18}. Each of the operators also has the probability of being applied to a given step, with $100\%$ probability meaning that the mutation is applied to every step of the agent's training. 

\begin{itemize}

    \item \textit{Reward Noise (RN)}: Adds a (Gaussian) noise to the true reward that the agent was meant to receive and returns it to the agent. 
    
    
    \item \textit{Mangled (M)}: Damages the correlation between collected experiences. This operator returns a random $s_{t+1}$ and $r_t$ which are not the state and reward the agent should receive according to its current state and the taken action. 
    
    \item \textit{Random (Ra)}: Similar to the mangled mutation operator, the \textit{Ra} mutation returns a $(s_t, a_t, r'_t, s'_{t+1})$ tuple to the agent. However, unlike the \textit{M} operator where $s'_{t+1}$ and $r'_t$ are selected randomly and are not associated with each other, the random operator returns some $s'_{t+1}$ and $r'_t$ to the agent which were sampled from the same experience tuples but have no association with $s_t$ and $a_t$.
    
    \item \textit{Repeat (R)}: Returns the previous observation to the agent. If the agent has two consecutive experiences in the form of $(s_t, a_t, r_t, s_{t+1})$ and $(s_{t+1}, a_{t+1}, r_{t+1}, s_{t+2})$, this operator returns $(s_{t+1}, a_{t+1}, r_t, s_{t+1})$. 
    \end{itemize}

\textbf{Agent-level:} As shown in \cite{nikanjam2022faults} many of the issues faced by developers in implementing RL algorithms, are the result of incorrect coding of RL concepts. The agent-level mutations are meant to simulate the faults that can happen when a developer makes mistakes in implementing the concepts of an RL-based agent in code. 
\begin{itemize}

    \item \textit{No Discount Factor (NDF)}: Removes the discount factor $\gamma$ from the reward calculation. 
    
    \item \textit{Missing Terminal State (MTS)}: Removes the terminal state of an episode (\ie reaching the goal, or falling into a trap...). 
    
    \item \textit{No Reverse (NR)}: Wrongly reverses the order of the received rewards and therefore makes the agent learn an incorrect association between the experiences. 
    
    \item \textit{Missing State Update (MSU)}: Removes the state update after the agent takes an action, meaning the agent will always see the same state in the experience tuple \eg $(s_t, a_{t+1}, r_{t+1}, s_{t+2})$ during training. 
    
    \item \textit{Incorrect Loss Function (ILF)}: Modifies the loss function used in the neural network that learns the policy. 

\end{itemize}

\textbf{Policy-level:} This category contains mutations affecting the policy of the agent. In general, neural networks being used, these mutations are similar to mutations that were previously defined in Supervised Learning \cite{humbatova2021deepcrime}.
\begin{itemize}
    \item \textit{Policy Activation Change (PAC)}: Changes the default activation function used in the policy network of the agent. 
    \item \textit{Policy Optimizer Change (POC)}: Modifies the default optimizer of the algorithm while keeping the original learning rate. 
\end{itemize}

\subsection{High Order Mutations}\label{sec:hom}

Based on our previously defined FOMs, we set to evaluate HOM in RL as well. To identify interesting HOMs, we follow a similar procedure to what was introduced in Jia et al. \cite{Jia09}, presented in Algorithm \ref{algo}.

\SetAlgoSkip{SkipBeforeAndAfter}
\begin{algorithm}[ht]
\scriptsize
 \SetAlgoLined
 \SetKwData{Left}{left}\SetKwData{This}{this}\SetKwData{Up}{up}
 \SetKwFunction{GenerateBoundsEnvironments}{GenerateBoundsEnvironments}
 \SetKwFunction{IsDifferent}{IsDifferent}
 \SetKwInOut{Input}{Input}\SetKwInOut{Output}{Output}
 \Input{$m$ FOM $\mathcal{F} = FOM_1, ..., FOM_m$, $n$ healthy agents $\mathcal{A} = A_1, ..., A_n$, for each FOM the relevant mutated agents $\mathcal{A}_{FOM_i} = A_{1, FOM_i}, ..., A_{n, FOM_i}$, initial environment $E_0$, parameters to modify $params$}
 \Output{The set of FOMs $\mathcal{F^*}$ to consider for the HOM}
 $\mathcal{E} = \{E_0, ..., E_p\}$ $\leftarrow$
 \GenerateBoundsEnvironments($\mathcal{A}, params, E_0$)\;
 $\mathcal{F^*}$ $\leftarrow$ $\emptyset$\;
 \ForEach{$f \in \mathcal{F}$}{
         i $\leftarrow$ 0\;
         \ForEach{$e \in \mathcal{E}$}
         {
            \uIf{\IsDifferent($\mathcal{A}, e, \mathcal{A}_f, e$)}{
                i++\;
            }
         }
         \uIf{i $\neq |\mathcal{E}|$ and i $\neq 0$}{
                $\mathcal{F^*}$ $\leftarrow$ $\mathcal{F^*} \cup f$
            }
 }
\caption{HOM generation algorithm}\label{algo}
\end{algorithm}

Starting from Line 2, the algorithm describes how we implement in our case the heuristic of \cite{Jia09} to determine which FOM to be considered for the HOM. We aim to determine which FOM are not trivial, \ie not killed by all test environments or by none. To evaluate if a mutation is killed by an environment, we use a distribution test over the rewards of healthy agents $\mathcal{A}$ evaluated on environment $e$ and mutated agents $\mathcal{A}_f$ evaluated on the same environment (function $IsDifferent$ in Algorithm \ref{algo}), using, for instance, the distribution test mentioned in Section \ref{sec:background}. However, to apply the heuristic, different test environments are needed. As such, we need a way to simply and effectively generate more test environments.

\subsection{Generating simple test environment for Mutation Testing}\label{sec:mutation_op:gen}


One way to generate new test environments in a relatively straightforward way that can be automated is to modify certain properties (\ie parameters) of said environments. As such, we define a test environment $E_i$ to be dependent on its physical parameters. For instance, the CartPole environment \cite{CartPoleURL} consists of a cart with a pole connected to it through a pivot. Two parameters of the environment are the cart's and pole's masses, which can be varied and then influence the agent's decision which in turn can reveal faults.

Finding diverse and non-trivial test cases which can have a different impact on the FOM is not obvious. Indeed, an exhaustive search is not practical because of the high number of parameter combinations and the computational cost of evaluating the agent's behavior, mutated or not, in each candidate test environment. At the same time, a random search might lead to many trivial test environments and similarly can be computationally expensive. Thus, leveraging some form of heuristic, even simplistic, is needed. Intuitively, environments close to the initial one have a high chance of yielding the same decision concerning the mutation, which would limit the relevance of such environments. At the same time, going too far away from the initial environment might unexpectedly affect the behavior of any agent and will also lower the relevance of the test environment. Consequently, finding the right balance between the two extremes is needed. This, however, assumes a certain continuity of the parameter space. Nonetheless, other works using a similar method such as Biagiola et al. \cite{Biagiola22} suggest such an approach can be used.

\begin{algorithm}[ht]
\scriptsize
 \SetAlgoLined
 \SetKwData{Left}{left}\SetKwData{This}{this}\SetKwData{Up}{up}
 \SetKwFunction{IsDifferent}{IsDifferent}
 \SetKwFunction{CheckPrecision}{CheckPrecision}
 \SetKwInOut{Input}{Input}\SetKwInOut{Output}{Output}
 \Input{$n$ healthy agents $\mathcal{A} = A_1, ..., A_n$, initial environment $E_0$, parameters to modify $params$}
 \Output{The set of boundary environments $\mathcal{E}$}
 $\mathcal{E} \leftarrow \{\}$\;
 \ForEach{$p \in params$}{ \tcp{Test environments on the upper boundaries}
         \uIf{$E_0.p \neq p.l_{upper}$}
         {
             $E_c \leftarrow E_0$\;
             $E_c.p \leftarrow p.l_{upper}$\;
             \uIf{not \IsDifferent($\mathcal{A}$, $E_c$, $\mathcal{A}$, $E_0$)}
             {
                $\mathcal{E} \leftarrow \mathcal{E} \cup E_c$\;
             }
             \Else
             {
                $E_b \leftarrow E_0$\;
                \While{\CheckPrecision($E_c$, $E_b$)}
                {
                   $p_m \leftarrow \frac{E_b.p + E_c.p}{2}$\;
                   \uIf{not \IsDifferent($\mathcal{A}$, $E_c$, $\mathcal{A}$, $E_0$)}
                   {
                        $E_b.p = p_m$\;
                   }
                   \Else
                   {
                        $E_c.p = p_m$\;
                   }
                }
                $\mathcal{E} \leftarrow \mathcal{E} \cup E_b$\;
             }
         }\Else
         {
             $\mathcal{E} \leftarrow \mathcal{E} \cup E_0$\;
         }
 }
...\;
\tcc{Similar for lower boundaries}
...\;

\For{$i \gets1$ \KwTo depth}{
    $\mathcal{E}_m \leftarrow \{\}$\;
    \For{$j \gets1$ \KwTo $|\mathcal{E}|$-1}{
        \uIf{$\mathcal{E}[j] \neq E_0$ AND $\mathcal{E}[j+1] \neq E_0$}
        {
            $E_c \leftarrow \frac{\mathcal{E}[j].p + \mathcal{E}[j+1].p}{2}$\;
            \uIf{not \IsDifferent($\mathcal{A}$, $E_c$, $\mathcal{A}$, $E_0$)}
             {
                $\mathcal{E_m} \leftarrow \mathcal{E}_m \cup \{\mathcal{E}[j], E_c, \mathcal{E}[j+1]\}$\;
             }
             \Else
             {
                $E_b \leftarrow E_0$\;
                \While{\CheckPrecision($E_c$, $E_b$)}
                {
                   $p_m \leftarrow \frac{E_b.p + E_c.p}{2}$\;
                   \uIf{not \IsDifferent($\mathcal{A}$, $E_c$, $\mathcal{A}$, $E_0$)}
                   {
                        $E_b.p = p_m$\;
                   }
                   \Else
                   {
                        $E_c.p = p_m$\;
                   }
                }
                $\mathcal{E}_m \leftarrow \mathcal{E}_m \cup \{\mathcal{E}[j], E_b, \mathcal{E}[j+1]\}$\;
            }
        }
    }
    $\mathcal{E} \leftarrow \mathcal{E}_m$\;
}
\uIf{$E_0 \notin \mathcal{E}$}
{
  $\mathcal{E} \leftarrow \mathcal{E} \cup E_0$\;
}
\caption{Generate Bounds Environments}\label{algo:gen}
\end{algorithm}

Therefore, we propose the following: by using only healthy agents, to decrease the computational cost, we can look for \enquote{frontier} environments, that is, environments that are different enough from the initial environment in terms of behavior but not too different. One way to find such a tipping point is to leverage binary search between healthy agents' reward on the initial environment and agents' reward on the generated candidate test environment, comparing them with the previously defined mutation killing definition (see Section \ref{sec:hom}). This is the goal of the function $GenerateBoundsEnvironments$ on Line 1 of Algorithm \ref{algo} and presented in Algorithm \ref{algo:gen}.

The method first looks for test environments along the parameter axes by applying a binary search over only one parameter between the initial environment ($E_0$) parameters and the defined search limits of the parameter $p.l$ (Line 2-25 Algorithm \ref{algo:gen}). In each step of the search, for the healthy agent, the distribution of reward obtained on environment $E_c$ is then tested against the distribution obtained on the initial environment $E_0$ (function $IsDifferent$ in Algorithm \ref{algo:gen}). The search stops when a certain precision is reached between the lower/upper boundaries of the environment's parameters. Then (Line 26-46), the algorithm will loop for a certain number of pre-defined \textit{depths}. It searches for test environments \textit{in-between} the environments that were calculated previously by iteratively applying the same method. Note that the algorithm was implemented for a 2-D space, since it is what was used in our experiments, yet it can be easily extended without loss of generality to $n$-D. Thus, for the 2-D case, from $4$ points, the algorithm will yield $8$ after the first \textit{depth}, and $16$ after the second. In the end, the set of test environments is returned in Algorithm \ref{algo}.

\section{Experimental Design and Results}\label{sec:res}

In this section, we introduce our research questions and describe the implementation of studied environments. Furthermore, we describe our RL training process, experimental design, mutation killing definitions used, and obtained results.
\subsection{Research questions}
To assess the effectiveness of Mutation Testing for RL, we formulate the following research questions (RQ):
\begin{itemize}
    \item[\textbf{RQ1:}] \textit{\rqone}
    \item[\textbf{RQ2:}] \textit{\rqtwo}
    \item[\textbf{RQ3:}] \textit{\rqthree}
\end{itemize}

\subsubsection{RQ1. Limitations of current mutation killing definitions}
As we detailed in Section \ref{sec:background}, Mutation Testing for RL can be applied in, at least, two ways. The first definition is proposed in \cite{Lu2022-RLmutation}, where a mutation is considered killed if the ratio of the average rewards over \textit{n} episodes gained by the healthy agent to the average of the mutated one is lower than a threshold $\theta$. The second definition is a distribution-wise statistical test as leveraged for instance in DeepCrime \cite{humbatova2021deepcrime}, which in RL's case can be based on the reward over \textit{n} episodes instead of the accuracy over the test set. However, the first definition is limited because of the stochasticity of RL while the second one is a straightforward adaptation of a Mutation Testing application for Supervised Learning, which might not be completely valid for RL as the reward is not necessarily equal to accuracy. \textbf{This question aims to evaluate the relevance of both those mutation killing definitions in RL over the mutation operators we defined previously.}

To have a fair comparison between the two definitions, we will need to modify the first approach (ratio of the average rewards over \textit{n} episodes gained by the healthy agent to the average of the mutated one) as it does not account for $N$ multiple agents. We will count the number of times the ratio is lower than a certain threshold among $N$ agents trained independently. The choice of the threshold $\theta$ can have an impact on the number of mutations found, as the higher the threshold is, the more likely it is that a mutation can be found. Yet at the same time, a too high $\theta$ might reduce the usefulness of the ratio in the case a mutated agent behaves similarly to a healthy agent in the test environment, e.g. an agent recovering from the mutation. Lu et al. \cite{Lu2022-RLmutation} chose a threshold of $\theta = 0.8$. We instead chose a more conservative value of $0.9$, which would make the test more likely to find a mutation. We will consider the mutation killed if at least $80\%$ of the $N$ ratios are lower than the $0.9$, which is roughly the power of the statistical test used for the distribution-wise test. The second method will use a statistical test over the reward of $N$ agents similar to the original implementation in DeepCrime \cite{humbatova2021deepcrime}.

\subsubsection{RQ2. Behaviors of FOM on generated test environments} The goal of this RQ is to investigate the behaviors of FOM with regard to the type of mutations used and the generated test environments. Test environments were generated following the procedure detailed in Section \ref{sec:mutation_op:gen} with a depth of $1$. Depth was chosen to keep a low number of environments for computation not to be too expensive. Increasing it would increase the number of environments generated and so, likely, the potential number of relevant FOM at the cost of increased processing time. The procedure allows us to \textbf{obtain a finer grain analysis of the FOM operators} by getting the number of test environments killing a given FOM. Moreover, it is then possible to analyze the parameters of the test environments in order \textbf{to assess what parameters' set is more likely to trigger a certain mutation} (for instance, for CartPole, if some mutations are more likely to be triggered when the mass of the cart is lowered or not). Finally, we will leverage FOM as presented in Section \ref{sec:mutation_op:gen}, to \textbf{deduce the interesting FOM to generate potential subsuming HOM}.

The test environments will be generated by altering two parameters of each environment: for CartPole, the \textit{mass of the cart and of the pole} will be modified, while for LunarLander, the \textit{gravity} and \textit{the side engine power} of the spacecraft will be modified. We refer the readers to our replication package \cite{rep-package} for the initial parameters as well as the search boundaries used.

\subsubsection{RQ3. Properties of generated HOM} Finally, similar to RQ2, in this RQ we aim to \textbf{analyze the HOM generated} in the previous experiments, by reusing our test environments. The goal is to understand which property the said generated HOM possesses, following the description briefly presented in Section \ref{sec:background:hom}. In particular, \textbf{we aim to see if we can generate Subsuming HOM} from our defined FOM and test environments.

Using chosen FOM from RQ2, we generate HOM that will in turn be used to train $N$ agents for each environment/algorithm/mutation operator. We will then evaluate them on the previously generated test environments and compare obtained results with those obtained from RQ2's FOM to deduce their properties.

\subsection{Implementation and models}
We used Python 3.8 and Stable Baselines 3 \cite{stable-baselines3} framework version 1.6.2 as a basis to implement our RL models and mutations. The motivation to use this framework is double: first, it allows us to evaluate some mutations that could directly affect the users of such a framework, for instance, a wrong activation or optimizer for the policy network. Secondly, it serves as a solid basis to implement mutations affecting the potential customized code of the user. Indeed, mutation can be introduced this way by overriding the base code by modifying only the relevant part, which ensures our code is less prone to unintended errors and allows better control over the mutation implementation.

To test our mutations, we chose two well-known environments in the RL community: \textit{CartPole} and \textit{LunarLander}. The CartPole environment \cite{CartPoleURL} consists of a cart with a pole connected to it through a pivot. The goal of the agent is to stabilize the pole and prevent it from falling by applying an appropriate amount of horizontal force by moving the cart left or right. LunarLander is a more complex environment that represents a spaceship that must land on a surface delimited by two flags. The agent can control the spaceship by throttling it in three directions (left, right, and down) \cite{LunarLander}.

We used three deep RL algorithms: similarly to the previous paper \cite{Lu2022-RLmutation}, we use Deep Q-Network (DQN) \cite{mnih2015human}. On top of that, instead of using plain Q-learning, which is a relatively simplistic algorithm, we will consider two other algorithms namely Advantage Actor-Critics (A2C) and Proximal Policy Optimization (PPO), which are classical algorithms in deep RL. This allows us to compare \textit{Off-Policy} algorithms (DQN) with \textit{On-Policy} algorithms (PPO, A2C), which are two major approaches in RL. We trained for each environment-algorithm-mutation, $N = 20$ agents with different seeds. We report for each algorithm/environment the average accumulated reward and standard deviation across the agents in Table \ref{tab:perf}. We also remind the readers of the mutations used in Table \ref{table:mutations_summary_table}.

\begin{table}[]
    \renewcommand{\arraystretch}{1.2}
    \centering
        \caption{Average rewards across the $20$ agents with the standard deviation in parenthesis.}
    \begin{tabular}{|c|c|c|c|}
          \hline
         & \textbf{PPO} & \textbf{A2C} & \textbf{DQN}  \\
         \hline
         \textbf{CartPole} & 500 (0) & 500 (0) & 414 (143)\\
         \hline
         \textbf{LunarLander} & 262 (16) & 141 (56) & 155 (84)\\
         \hline
    \end{tabular}
    \label{tab:perf}
    \vspace{-2em}
\end{table}

Some mutations are reliant on some parameters and so they will be defined with both the mutation operator identifier and the parameter. For instance, \textit{Policy Activation Change} requires defining which new activation will be used, such as Stochastic Gradient Descent (SGD) and so will be noted as $PAC\_SGD$. Environment-level mutations are based on some probability of being applied for each given step, and so $M\_1.0$ means that the mutation operator \textit{Mangled} will be used with a $100\%$ probability at each step. If no parameters are needed, then just the identifier is used, for instance, \textit{No Reverse} will simply be noted as $NR$. All mutations are used for all agents, except \textit{NR} and \textit{PAC-ReLU} for \textit{DQN}, the first one not applying to \textit{DQN} while the second one being the default activation of \textit{DQN}.


\begin{table}[htb]
\caption{Mutation operators summary}

\resizebox{\columnwidth}{!}
    {\begin{tabular}{m{2cm} p{3.75cm} p{4.5cm}}
        \toprule
        Category & Operator & Description \\ 
        \midrule
            \multirow{10}{*}{Environment-level} & Reward noise (RN) & Adding noise to the reward the agent receives \\
            \rule{0pt}{4ex}
            & Mangled (M) & Returning next state and reward which are not related to each other \\
            \rule{0pt}{4ex}
            & Random (Ra) & Returning next state and reward that are related to each other but not related to the action taken \\
            \rule{0pt}{4ex}
            & Repeat (R) & Returning next state and reward from previous observation \\ 
        \midrule
            \multirow{12}{*}{Agent-level} & No discount factor (NDF) & No discount factor during calculating cumulative rewards \\
            \rule{0pt}{4ex}
            & No reverse (NR) & Not reversing the order of the received rewards during calculating cumulative rewards \\
            \rule{0pt}{4ex}
            & Missing state update (MSU) & Not updating agent's observations \\
            \rule{0pt}{4ex}
            & Missing terminal state (MTS) & Failing to save the terminal state observation \\
            \rule{0pt}{4ex}
            & Incorrect loss function (ILF) & Defining an incorrect loss function (wrong formula, etc.) \\
            
        \midrule
            \multirow{3}{*}{Policy-level} & Policy activation change (PAC) & Different activation for agent's neural network \\
            \rule{0pt}{4ex}
            & Policy optimizer change (POC) & Different optimizer for agent's neural network \\ 
        \bottomrule
    \end{tabular}}
\label{table:mutations_summary_table}
\vspace{-2em}
\end{table}

\subsection{Results}

\begin{table*}[!ht]
\renewcommand{\arraystretch}{1.2}
\caption{FOM mutation test results. $\checkmark$ means mutation is killed while ~\xmark~ means mutation is not killed or the statistical power of the test is too low (inconclusive). The killing criteria used are \textbf{AVG}: Average Reward Mutant/Healthy with the value in parentheses being the proportion of ratios below the threshold $\theta$, \textbf{R}: Reward-based statistical test, and \textbf{DtR}: Distance to Healthy Reward statistical test. \enquote{\textbf{-}} means mutation is not applicable.}
\begin{center}
\resizebox{\textwidth}{!}{\begin{tabular}{|c|c|c|c|c|c|c|c|c|c|c|c|c|c|c|}
\hline
\multirow{2}{*}{\textbf{Environment}}&\textbf{DRL}&\textbf{Killing}& \multicolumn{12}{c|}{\textbf{Mutations}}\\ \cline{4-15}
& \textbf{Algorithm}& \textbf{Criteria} & \textbf{ILF} & \textbf{M-1.0} & \textbf{R-1.0} & \textbf{Ra-1.0} & \textbf{RN-1.0} & \textbf{NDF} & \textbf{NR} & \textbf{MSU} & \textbf{MTS} & \textbf{PAC-ReLU} & \textbf{PAC-Sigmoid} & \textbf{POC-SGD} \\
\hline
\multirow{9}{*}{CartPole}& \multirow{3}{*}{PPO} & AVG & $\checkmark$ (1.0) & $\checkmark$ (1.0) & \xmark~(0.0) & $\checkmark$ (0.95) & \xmark~(0.0) & \xmark~(0.65) & $\checkmark$ (1.0) & $\checkmark$ (1.0) & \xmark~(0.05) & \xmark~(0.05) & \xmark~(0.25) & $\checkmark$ (1.0) \\ \cline{3-15}

& & R & $\checkmark$ & $\checkmark$ & \xmark & $\checkmark$ & \xmark & $\checkmark$ & $\checkmark$ & $\checkmark$ & \xmark &\xmark & $\checkmark$ & $\checkmark$ \\ \cline{3-15}

& & DtR & $\checkmark$ & $\checkmark$ & \xmark & $\checkmark$ & \xmark & $\checkmark$ & $\checkmark$ & $\checkmark$ & $\checkmark$ & $\checkmark$ & $\checkmark$ & $\checkmark$ \\ \clineB{2-15}{3}

& \multirow{3}{*}{A2C} & AVG & $\checkmark$ (1.0) & $\checkmark$ (0.95) & \xmark~(0.15) & $\checkmark$ (1.0) & \xmark~(0.25) & \xmark~(0.35) & $\checkmark$ (0.9) & $\checkmark$ (1.0) & \xmark~(0.05) & $\checkmark$ (1.0) & \xmark~(0.2) & $\checkmark$ (1.0) \\ \cline{3-15}

& & R & $\checkmark$ & $\checkmark$ & \xmark & $\checkmark$ & $\checkmark$ & $\checkmark$ & $\checkmark$ & $\checkmark$ & \xmark & $\checkmark$ & $\checkmark$ & $\checkmark$ \\ \cline{3-15}

& & DtR & $\checkmark$ & $\checkmark$ & $\checkmark$ & $\checkmark$ & $\checkmark$ & $\checkmark$ & $\checkmark$ & $\checkmark$ & $\checkmark$ & $\checkmark$ & $\checkmark$ & $\checkmark$ \\ \clineB{2-15}{3}

& \multirow{3}{*}{DQN} & AVG & $\checkmark$ (1.0) & $\checkmark$ (1.0) & $\checkmark$ (0.9) & $\checkmark$ (1.0) & \xmark~(0.3) & $\checkmark$ (1.0) & - & $\checkmark$ (1.0) & $\checkmark$ (1.0) & - & $\checkmark$ (0.8) & $\checkmark$ (1.0) \\ \cline{3-15}

& & R & $\checkmark$ & $\checkmark$ & $\checkmark$ & $\checkmark$ & \xmark & $\checkmark$ & - & $\checkmark$ & $\checkmark$ & - & $\checkmark$ & $\checkmark$ \\ \cline{3-15}

& & DtR & $\checkmark$ & $\checkmark$ & $\checkmark$ & $\checkmark$ & \xmark & $\checkmark$ & - & $\checkmark$ & $\checkmark$ & - & $\checkmark$ & $\checkmark$ \\ \clineB{1-15}{3}
\hline

\multirow{9}{*}{LunarLander}& \multirow{3}{*}{PPO} & AVG & $\checkmark$ (1.0) & $\checkmark$ (1.0) & \xmark~(0.2) & $\checkmark$ (1.0) & \xmark~(0.05) & $\checkmark$ (0.95) & $\checkmark$ (1.0) & $\checkmark$ (1.0) & \xmark~(0.05) & \xmark~(0.05) & $\checkmark$ (0.95) & $\checkmark$ (1.0) \\ \cline{3-15}

& & R & $\checkmark$ & $\checkmark$ & \xmark & $\checkmark$ & \xmark & $\checkmark$ & $\checkmark$ & $\checkmark$ & \xmark &\xmark & $\checkmark$ & $\checkmark$ \\ \cline{3-15}

& & DtR & $\checkmark$ & $\checkmark$ & \xmark & $\checkmark$ & \xmark & $\checkmark$ & $\checkmark$ & $\checkmark$ & $\checkmark$ & \xmark & $\checkmark$ & $\checkmark$ \\ \clineB{2-15}{3}

& \multirow{3}{*}{A2C} & AVG & $\checkmark$ (1.0) & $\checkmark$ (1.0) & \xmark~(0.3) & $\checkmark$ (1.0) & \xmark~(0.45) & \xmark~(0.45) & $\checkmark$ (0.85) & $\checkmark$ (1.0) & \xmark~(0.45) & \xmark~(0.2) & \xmark~(0.6) & $\checkmark$ (1.0) \\ \cline{3-15}

& & R & $\checkmark$ & $\checkmark$ & \xmark & $\checkmark$ & \xmark & \xmark & $\checkmark$ & $\checkmark$ & \xmark & $\checkmark$ & \xmark & $\checkmark$ \\ \cline{3-15}

& & DtR & $\checkmark$ & $\checkmark$ & \xmark & $\checkmark$ & $\checkmark$ & $\checkmark$ & $\checkmark$ & $\checkmark$ & \xmark & $\checkmark$ & \xmark & $\checkmark$ \\ \clineB{2-15}{3}

& \multirow{3}{*}{DQN} & AVG & $\checkmark$ (0.95) & $\checkmark$ (0.95) & $\checkmark$ (0.9) & $\checkmark$ (0.95) & \xmark~(0.6) & $\checkmark$ (0.95) & - & $\checkmark$ (0.95) & \xmark~(0.7) & - & $\checkmark$ (0.8) & $\checkmark$ (0.95) \\ \cline{3-15}

& & R & $\checkmark$ & $\checkmark$ & $\checkmark$ & $\checkmark$ & \xmark & $\checkmark$ & - & $\checkmark$ & \xmark & - & \xmark & $\checkmark$ \\ \cline{3-15}

& & DtR & $\checkmark$ & $\checkmark$ & $\checkmark$ & $\checkmark$ & \xmark & $\checkmark$ & - & $\checkmark$ & \xmark & - & $\checkmark$ & $\checkmark$ \\ \cline{2-15}
\hline
\end{tabular}}
\label{fig:comp_Mutation Testing_func}
\end{center}
\vspace{-2em}
\end{table*}

\subsubsection{RQ1. Existing limitations of current mutation killing definitions}

Results of FOM for a given algorithm/environment and killing definition are given in Table \ref{fig:comp_Mutation Testing_func} with \textit{AVG} being the method using the average of the ratio and \textit{R} being the method using reward-wise statistical comparison. As one can see, the limitations of \textit{AVG} we briefly introduced in Section \ref{sec:background:Mutation Testing_ml} are shown: for multiple mutation operators, no matter the environment/algorithm, the agents evaluated by the ratio might yield a very different result. If some mutation operators such as \textit{ILF} or \textit{POC-SGD} seem to be killed by the test environment on all healthy/mutated pairs, many mutations show a relatively low number of pairs declaring them killed. It is even possible that, for mutations with relatively small effects, the mutated agent ends up with a higher reward than the healthy agent if the initialization was not favorable for a particular healthy agent. As such, it seems that the stochasticity of the training process greatly influences the ratio obtained, leading to mutation not being killed across multiple test runs. On the other hand, using a reward distribution statistical test (\textit{R}) leads to more mutations being killed in all cases except for \textit{LunarLander/DQN}. In particular, this method allows mutations to be killed while $AVG$ only had a handful of agents pair declaring said mutations killed using the same test environment. For instance, \textit{CartPole/A2C/RN-1.0} where only $25\%$ of ratios revealed the mutation.

\begin{tcolorbox}[colback=blue!5,colframe=blue!40!black]
\textbf{RQ1-1} Previously introduced mutation killing definition in RL based on a ratio between healthy/mutated agents is not sufficient as it might miss a high number of mutations because of the stochasticity of the training process, particularly for mutations with low effect on the training which are harder to find. Using distribution-wise statistical tests based on the reward improves this aspect.
\end{tcolorbox}

Nonetheless, for the same test environment, the distribution-wise test \textit{R} still does not allow the detection of a high number of mutations, even mutations for which there are a sizable amount of ratios declaring the mutation killed with the \textit{AVG} definition, such as \textit{LunarLander/A2C/NDF}. In those cases, while some mutated agents can exhibit a lower reward than some healthy agents, both healthy and mutated agents' reward distribution might not exhibit a statistically significant difference. This can particularly be the case when a few agents end up exhibiting a very different reward because of the effect of the mutation not being overcome by the training or being accentuated by a non-favorable initialization as initial seed can have a high impact on the training \cite{Henderson18, Agarwal21}. Removing such data points is not ideal as it masks potentially useful information. One way to account for those data is to leverage the fact that we know the variation between healthy/mutated agents but also \textit{between} healthy agents. And so any potential outlier would lead to a large difference between them. Such variation can be estimated by calculating the Hellinger distance \cite{cramer46} of the rewards between samples of the healthy agents' distribution (\textit{intra} distance) and the same distance between samples of the healthy/mutated agents distribution (\textit{inter} distance). The Hellinger distance is a metric bounded between 0 and 1, with 0 meaning both distributions are the same which thus can be interpreted as a measure of similarity between the distributions.
For the discrete case, it is defined for two distributions P and Q as:
\begin{equation}
    H(P, Q) = \frac{1}{\sqrt{2}} \lvert \lvert P - Q\rvert \rvert_2
\end{equation}
where $\lvert \lvert . \rvert \rvert_2$ is the $L^2$ norm.

By repeating the calculation by sampling from both agents' reward distribution, one can obtain the distributions of the \textit{inter}/\textit{intra} distance and use the same statistical test used in the definition \textit{R}. In practice, we will use samples of $10$ agents out of $20$ to calculate the distance. The results are presented in Table \ref{fig:comp_Mutation Testing_func} in the \textit{DtR} rows.

Using this approach improves on using simply the rewards of healthy/mutated agents. In particular, it manages to lead to some mutations with relatively low \textit{AVG} scores to be killed by the same test environment, such as \textit{LunarLander/PPO/MTS}, as \textit{DtR} allows to account for the previous potential outlier problem. Nonetheless, one can see the method is not perfect as mutations such as \textit{LunarLander/DQN/RN} end up not being killed while a sizable number of agent pairs are declared mutant by \textit{AVG}. In those cases, the variation among healthy agents might be too pronounced compared to the ones between healthy and mutated agents, which is why \textit{DtR} can not catch the mutation. This however highlights an important observation: the choice of the mutation killing definition design in RL is a crucial step to determine if a mutation is killed or not.

\begin{tcolorbox}[colback=blue!5,colframe=blue!40!black]
\textbf{RQ1-2} More than choosing which mutation operators to consider, careful selection of \textit{how} to use Mutation Testing when applied to RL is another important parameter. For instance, using Hellinger distance to healthy reward instead of plain reward allowed for more mutations being killed for the same test environment. Thus, investigating different mutation killing definition designs in RL is a crucial step.
\end{tcolorbox}

\subsubsection{RQ2. Behaviors of FOM on generated test environments} Following the procedure of Section \ref{sec:mutation_op:gen}, we generated several test environments for each algorithm/environment and made a number of observations. Results are presented in Table \ref{fig:test_gen_res}. Mutations flagged in green for a given environment/algorithm mean that the mutation is not trivial in that case and so will be relevant when we need to generate HOMs. Note that we focus only on distribution-wise tests from this point on, as RQ1 illustrated the drawback of the mutation killing definition in \cite{Lu2022-RLmutation}.

The first observation we make is that five mutations are killed by all generated test environments, namely \textit{ILF}, \textit{MSU}, \textit{NR}, \textit{POC-SGD}, \textit{M-1.0} and \textit{Ra-1.0}. Those mutations have too much of an impact not to be detected, which is also highlighted in Table \ref{fig:comp_Mutation Testing_func}, as those mutations yielded a score $> 0.85$ when the initial test environment was used with the ratio method (\textit{AVG}). In those cases, the stochasticity is masked by the high impact of the mutations and all trained agents behave similarly. Thus, generating multiple test environments allowed us to see that those mutations might not be much of interest. Indeed, they are rather trivial to detect and so can probably be ignored.

Secondly, we see the test environments generated on \textit{CartPole} are relatively more likely to catch mutations with most of the mutations being killed by at least half the test environments compared to the ones generated on \textit{LunarLander}. While the number of test environments generated might play a role, it is likely because \textit{LunarLander} is a more complex environment than \textit{CartPole}, so mutations might become harder to detect as the environment complexity increases. For instance, \textit{CartPole} seems to be less sensitive to the \textit{MTS} or \textit{PAC} mutations, \ie removing the terminal state of an episode or changing the activation function of the policy network does not seem to affect much the agents trained on \textit{CartPole} contrary to \textit{LunarLander}.

\begin{tcolorbox}[colback=blue!5,colframe=blue!40!black]
\textbf{RQ2-1} Contrary to using only the initial test environment, generating additional test environments allows us to roughly evaluate which mutations might be trivial and which are more interesting based on the number of environments killing them. We also found that it appears that, the more complex the environments the more likely the mutation is not to be found. 

\end{tcolorbox}

\begin{table*}
\caption{Number of test environments killing each FOM. Green Cells are FOM that will be used to generate HOM.}
\renewcommand{\arraystretch}{1.2}
\begin{center}
\resizebox{\textwidth}{!}{\begin{tabular}{|c|c|c|c|c|c|c|c|c|c|c|c|c|c|c|}
\hline
\multirow{2}{*}{\textbf{Environment}}&\textbf{DRL}&\textbf{Killing}& \multicolumn{12}{c|}{\textbf{Mutations}}\\ \cline{4-15}
& \textbf{Algorithm}& \textbf{Criteria} & \textbf{ILF} & \textbf{M-1.0} & \textbf{R-1.0} & \textbf{Ra-1.0} & \textbf{RN-1.0} & \textbf{NDF} & \textbf{NR} & \textbf{MSU} & \textbf{MTS} & \textbf{PAC-ReLU} & \textbf{PAC-Sigmoid} & \textbf{POC-SGD} \\
\hline
\multirow{6}{*}{CartPole}& \multirow{2}{*}{PPO} & R & 4/4 & 4/4 & 0/4 & 4/4 & \cellcolor{green!25}3/4 & 4/4 & 4/4 & 4/4 & \cellcolor{green!25}3/4 & \cellcolor{green!25}3/4 & 4/4 & 4/4 \\ \hhline{~|~|-|-|-|-|-|-|-|-|-|-|-|-|-|}
& & DtR & 4/4 & 4/4 & \cellcolor{green!25}3/4 & 4/4 & 4/4 & 4/4 & 4/4 & 4/4 & 4/4 & 4/4 & 4/4 & 4/4\\ \hhline{~|-|-|-|-|-|-|-|-|-|-|-|-|-|-|}

& \multirow{2}{*}{A2C} & R & 4/4 & 4/4 & 0/4 & 4/4 & 4/4 & \cellcolor{green!25}3/4 & 4/4 & 4/4 & \cellcolor{green!25}3/4 & 4/4 & \cellcolor{green!25}2/4 & 4/4 \\ \hhline{~|~|-|-|-|-|-|-|-|-|-|-|-|-|-|}
& & DtR & 4/4 & 4/4 & 4/4 & 4/4 & 4/4 & 4/4 & 4/4 & 4/4 & 4/4 & 4/4 & 4/4 & 4/4\\
\hhline{~|-|-|-|-|-|-|-|-|-|-|-|-|-|-|}

& \multirow{2}{*}{DQN} & R & 4/4 & 4/4 & 4/4 & 4/4 & 0/4 & 4/4 & - & 4/4 & 4/4 & - & 4/4 & 4/4 \\ \hhline{~|~|-|-|-|-|-|-|-|-|-|-|-|-|-|}
& & DtR & 4/4 & 4/4 & 4/4 & 4/4 & \cellcolor{green!25}3/4 & 4/4 & - & 4/4 & 4/4 & - & 4/4 & 4/4  \\
\hline

\multirow{6}{*}{LunarLander}& \multirow{2}{*}{PPO} & R & 8/8 & 8/8 & 0/8 & 8/8 & \cellcolor{green!25}1/8 & 8/8 & 8/8 & 8/8 & 0/8 & \cellcolor{green!25}4/8 & 8/8 & 8/8 \\ \hhline{~|~|-|-|-|-|-|-|-|-|-|-|-|-|-|}
& & DtR & 6/8 & 6/8 & \cellcolor{green!25}2/8 & 6/8 & \cellcolor{green!25}3/8 & 6/8 & 6/6 & 6/6 & \cellcolor{green!25}3/6 & \cellcolor{green!25}3/6 & 6/6 & 6/6 \\ \hhline{~|-|-|-|-|-|-|-|-|-|-|-|-|-|-|}

& \multirow{2}{*}{A2C} & R & 9/9 & 9/9 & 0/9 & 9/9 & 0/9 & \cellcolor{green!25}6/9 & 9/9 & 9/9 & \cellcolor{green!25}2/9 & \cellcolor{green!25}3/9 & \cellcolor{green!25}5/9 & 9/9 \\ \hhline{~|~|-|-|-|-|-|-|-|-|-|-|-|-|-|}
& & DtR & 9/9 & 9/9 & \cellcolor{green!25}2/9 & 9/9 & \cellcolor{green!25}4/9 & \cellcolor{green!25}7/9 & 9/9 & 9/9 & \cellcolor{green!25}4/9 & \cellcolor{green!25}7/9 & \cellcolor{green!25}6/9 & 9/9 \\
\hhline{~|-|-|-|-|-|-|-|-|-|-|-|-|-|-|}

& \multirow{2}{*}{DQN} & R & 9/9 & 9/9 & \cellcolor{green!25}8/9 & 9/9 & 0/9 & 9/9 & - & 9/9 & 0/9 & - & 0/9 & 9/9 \\ \hhline{~|~|-|-|-|-|-|-|-|-|-|-|-|-|-|}
& & DtR & 9/9 & 9/9 & 9/9 & 9/9 & \cellcolor{green!25}4/9 & 9/9 & - & 9/9 & \cellcolor{green!25}6/9 & - & \cellcolor{green!25}4/9 & 9/9 \\
\hline
\end{tabular}}
\label{fig:test_gen_res}
\end{center}
\vspace{-2em}
\end{table*}

In a second step, it is possible to go beyond the raw number of test environments killing a certain mutation and, instead, to inspect which test environment kills the mutation. Indeed, as we generated test environments through the modification of the initial environment parameters, this can allow us to shed some light on which parameter a mutation might be more easily sensitive to. By doing so, we can determine which test environments can help identify certain potential faults, thus leading to some sort of fault-detection method. Because of space constraints, we will report one example using \textit{R} and we refer the reader to our implementation repository \cite{rep-package} for all the raw results. In the case of \textit{LunarLander/PPO/PAC\_ReLU}, in Figure \ref{fig:exp_sep}, test environments with lower (absolute) gravity compared to the initial environment kill the mutation while the ones with higher (absolute) gravity do not. Thus, the mutation seems to be affected somehow by the gravity parameter. When the gravity is close to the one in the initial environment, higher engine power will be crucial to kill the mutation.

\begin{figure}
    \centering
    \includegraphics[width=\linewidth]{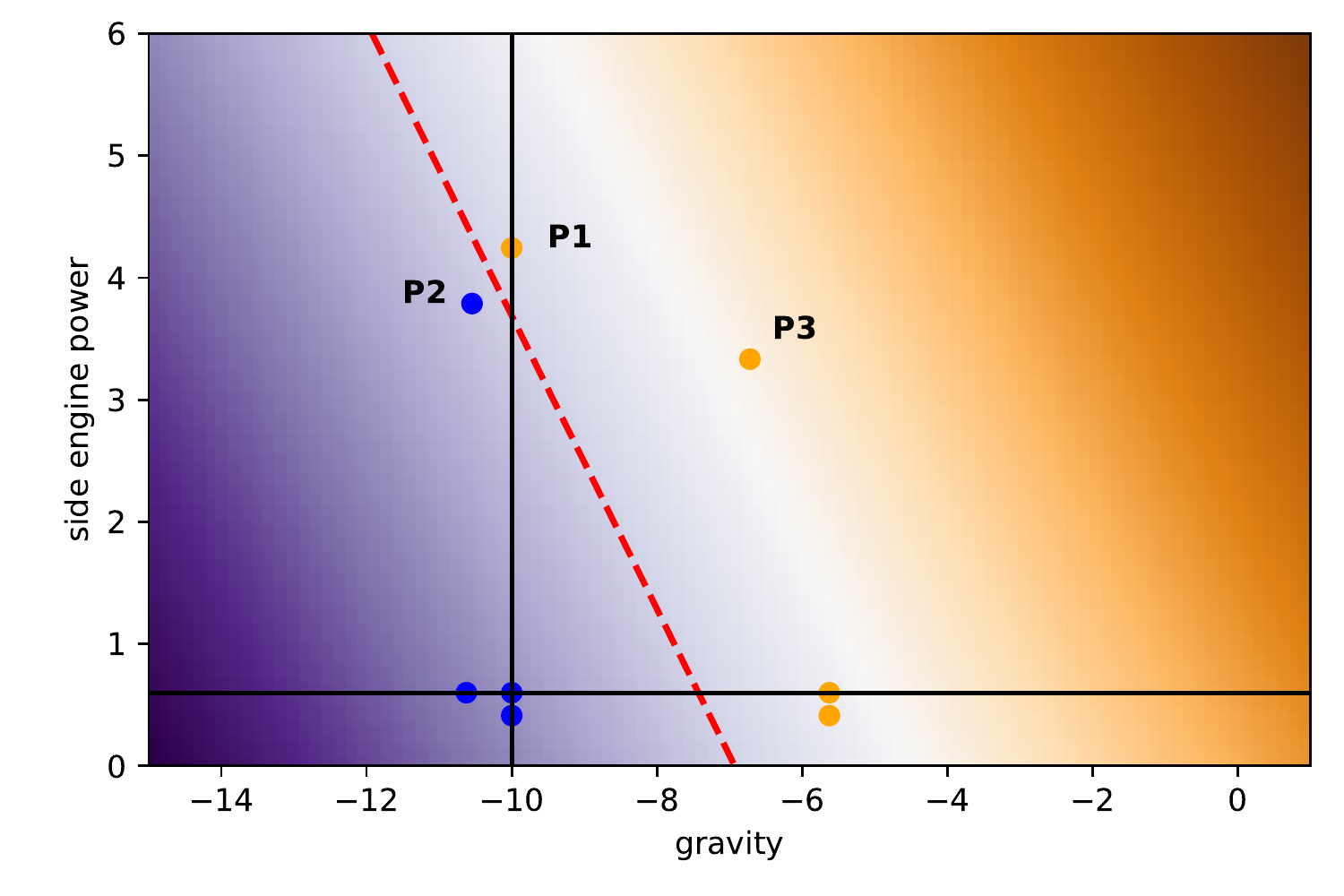}
    \caption{Generated test environments for \textit{LunarLander/PPO/PAC\_ReLU} and a potential way to separate them based on whether or not they killed the mutation. \textcolor{orange}{Orange} points kill the mutation while the \textcolor{blue}{Blue} ones don't. The origin is centered on the initial environments.}\label{fig:exp_sep}
    \vspace{-1.5em}
\end{figure}

Test environments are depicted as orange/blue dots, depending on whether or not they kill the mutation, and we can see they can be easily separated linearly following the previous description. In particular, we can see that both test environments $P1$ and $P2$ seem to be close to some frontiers for this mutation since, while being close in parameters space, they lead to an opposite decision on the mutation. Note that the red frontier is arbitrary as we do not have access to the exact frontier, as it would be potentially too computationally expensive to find it as we explained in Section \ref{sec:mutation_op:gen}. Thus, it could be possible to find environments between $P3$ and the initial environment that could still kill the mutation. We just know $P3$ is an environment for which the healthy agents' reward distribution is at the limit of being different from the distribution observed in the initial environment, but it gives no information on the distribution of the mutated agents. While it might not be possible to draw meaningful information for all mutations, especially on a reduced set of test environments, it shows nonetheless that generating test environments in that way also allows us to explore a potential link between parameters of the environments and their impact on the mutation.

\begin{tcolorbox}[colback=blue!5,colframe=blue!40!black]
\textbf{RQ2-2} By mapping, for a given mutation, which generated test environments kill it or not, we can analyze which of the parameters of the test environments affect the decision to kill the mutation. This outlines some form of fault-detection method.
\end{tcolorbox}

\subsubsection{RQ3. Properties of generated HOM} Following RQ2, we gathered for each environment/algorithms/killing definition the non-trivial FOM (\ie nor killed by all test environments nor killed by none), see Table \ref{fig:test_gen_res}. To generate HOM, we need at least two FOM as we stick to HOM of order 2. We then trained new mutated agents based on the gathered HOM in the same way as FOM. Finally, mutated agents were evaluated using the previously generated test environments depending on the mutation killing definition (\textit{R} or \textit{DtR}), and the type of HOM was analyzed based on the classification we introduced in Section \ref{sec:background:hom}. Results are presented in Table \ref{tab:hom_types}.

\begin{table}[]
\renewcommand{\arraystretch}{1.2}
\caption{Types of HOM generated using RQ2 non-trivial FOM. If no HOM was generated, then a \enquote{-} is used. \textit{\textbf{NS}}: Non Subsuming, \textit{\textbf{WSC}}: Weakly Subsuming Coupled, \textit{\textbf{WSD}}: Weakly Subsuming Decoupled, \textit{\textbf{SSC}}: Strongly Subsuming Coupled.}
\centering
\resizebox{\columnwidth}{!}{
\begin{tabular}{|c|c|c|c|c|c|c|c|c|}
\hline
\multirow{2}{*}{\textbf{Environment}}&{\textbf{DRL}}&{\textbf{Killing}}& \multicolumn{5}{c|}{\textbf{HOM types}}\\ \cline{4-8}
& \textbf{Algorithm}& \textbf{Criteria}& \textbf{HOM} & \textbf{NS} & \textbf{WSC} & \textbf{WSD} & \textbf{SSC}\\
\hline
\multirow{6}{*}{CartPole}& \multirow{2}{*}{PPO} & R & 3 & 1 & 0 & 0 & 2 \\ \cline{3-8}

& & DtR & - & - & - & - & - \\ \clineB{2-8}{3}

& \multirow{2}{*}{A2C} & R & 3 & 2 & 1 & 0 & 0 \\ \cline{3-8}

& & DtR & - & - & - & - & - \\ \clineB{2-8}{3}

& \multirow{2}{*}{DQN} & R & - & - & - & - & - \\ \cline{3-8}

& & DtR & - & - & - & - & - \\ \clineB{1-8}{3}
\hline

\multirow{6}{*}{LunarLander}& \multirow{2}{*}{PPO} & R & 1 & 1 & 0 & 0 & 0 \\ \cline{3-8}

& & DtR & 6 & 1 & 5 & 0 & 0 \\ \clineB{2-8}{3}

& \multirow{2}{*}{A2C} & R & 5 & 2 & 3 & 0 & 0 \\ \cline{3-8}

& & DtR & 14 & 6 & 8 & 0 & 0 \\ \clineB{2-8}{3}

& \multirow{2}{*}{DQN} & R & - & - & - & - & - \\ \cline{3-8}

& & DtR & 3 & 0 & 3 & 0 & 0 \\ \cline{2-8}
\hline
\end{tabular}}
\label{tab:hom_types}
\vspace{-2.3em}
\end{table}

As we can see, following the procedure mentioned for RQ2, we do not end up with a high number of non-trivial FOM and so the pool of generated HOM is relatively small with even some configurations not yielding any. Nonetheless, we managed to obtain $12$ HOM when the \textit{R} mutation killing definition was used and $23$ with \textit{DtR} but only in \textit{LunarLander}. Among those, Non-Subsuming (\textit{NS}) constitutes $50\%$ of HOM using \textit{R} method but only $30\%$ using \textit{DtR}, the difference between the two methods potentially being explained by the increased sensitivity of \textit{DtR}. The remaining Subsuming HOM are mostly of the \textit{Weakly Subsuming Coupled} (WSC) types and generally compose more than half the generated HOM for each configuration, which is similar to results obtained by Jia et al \cite{Jia09} with HOM in some traditional software programs. Interestingly, we found 2 HOM that fit the type \textit{Strongly Subsuming Coupled} (SSC), that is, HOM for which test environments killing said HOM also kill its constituent FOM. Aside from those two, no other \textit{SSC} and no \textit{WSD} were found. Even in Jia et al. case, \textit{SSC} represents a rare occurrence ($< 1\%$ of Subsuming HOM) and so the fact we found none is not too surprising judging by our limited number of HOM/test environments. Nonetheless, we showed that Subsuming HOM (even if \textit{Weakly} ones), which are more complex and subtle mutants, could be generated and make up for more than half the HOM generated.

\begin{tcolorbox}[colback=blue!5,colframe=blue!40!black]
\textbf{RQ3} HOM generated from non-trivial FOM, while few, are in the majority Subsuming HOM, which de facto makes them more interesting cases to use as we pointed out in Section \ref{sec:background:hom}. If more subtle Subsuming HOM such as \textit{SSC} are not as widely represented, the fact that some can be generated shows that such property is reachable in RL too.
\end{tcolorbox}

\section{Threats to Validity}\label{sec:threats}
\textbf{Construct validity:} The design choices of Mutation Testing could affect our results. Since our main goal was not to design a new mutation killing definition for Deep Learning but rather to adapt and assess how existing approaches fared, all the mutation killing definitions used are based on existing approaches. Moreover, while the hyper-parameters used could play a role in declaring a mutation killed (\textit{p-value}, threshold \textit{$\theta$}...), we preferred to stick to the hyper-parameters given in each original implementation of the killing definitions and leave to future work the study of the influence of hyper-parameters. The way we searched the parameters space to generate test environments could also have impacted our results. Nonetheless, the goal here was simply to get a simple and effective approach that could be automated and serve as a basic heuristic for future work. The rest of our methodology, such as the definition of the properties of HOM, is grounded in the scientific literature and is taken as they were originally defined.
 
\textbf{Internal validity:} Mutation operators chosen could affect our results. While we can not be exhaustive, we made sure to create operators based on existing faults or taxonomy and to provide sufficient diversity in terms of the effect on the code (\textit{agent}-based, \textit{environment}-based, \textit{policy}-based). For the \textit{environment-based} operators, while the probability of application of $1.0$ might not be real to simulate sensor defaults, it was necessary to avoid potential bias over which step would be affected by the mutation when the probability is set to $< 1.0$. The environment parameters modified to generate the test environments could also impact the results. However, the goal was to show we could generate automatically simple and effective test environments that could be leveraged to analyze Mutation Testing in RL. As such, the choice of the parameters on its own is not as relevant.
 
\textbf{External validity:} The choice of environment types/algorithms could impact the generalization of our experiments. We made sure to select two environments as well as three algorithms that are generally used as a benchmark in RL.  
 
\textbf{Reliability validity:} Implementing the mutations could lead to some unintended faults. We used the Stable-Baselines framework as a basis to implement our mutations to lower the risk. We also make our implementation and artifacts available \cite{rep-package}.

\section{Related work}\label{sec:rel}
Gur et al. \cite{gur2021environment} proposed an agent which produces environments depending on the learning agent's level of skill. Their approach allows for more methodological training of agents and increases their robustness and ability to generalize. While interesting, their approach jointly trains the environment generator with the agents which is not doable within the Mutation Testing scope. To combat the problem of overfitting, Cobbe et al. \cite{cobbe2020leveraging} introduced the Procgen benchmark. This benchmark comprises 16 procedurally generated environments that allow practitioners to test the generalization of the agents. Their idea allows for testing agents yet the main focus of Mutation Testing is to evaluate test set quality. Nonetheless, it would be interesting in future work to evaluate \tool{} on their environments, as they are designed to push the generalization property of the algorithm and so might induce different behavior over the mutations. In another direction, a meta-heuristic-based algorithm could be another venue to improve the generation of relevant test environments \cite{Humeniuk22}. Finally, Biagiola et al. \cite{Biagiola22} focused on generating test environments through a combination of binary/exponential search, to map the adaptation/anti-regression performance of an agent through the lens of continual learning. In our work, we similarly generate test environments by acting on their parameters, yet we do not test for the adaptation nor do we retrain the agent with continual learning. If anything, we similarly test for anti-regression of our healthy agents on both initial/generated test environments. Nonetheless, their approach could be used to improve our test generation part.

As mentioned in Introduction and Section \ref{sec:background}, some frameworks applied Mutation Testing to Supervised Learning \cite{Ma18, Hu19, humbatova2021deepcrime}. While we reuse some of their mutation operators/design choices, we specifically tailored our approach for RL. Lu et al. \cite{Lu2022-RLmutation} also tackled Mutation Testing in RL and studied how well-crafted environments could be used to reveal a particular mutation which is more akin to the fault detection method. On the contrary, we focus on generating, in an automatic way, simple yet effective test environments that can be used to reveal more complex faults, \ie HOM. We also show their Mutation Testing design choice could be a limiting factor because of the stochasticity of RL. Shen et al. \cite{Shen21} used decision boundaries of Supervised Learning models in order to find a subset of the test set that is more likely to trigger the mutation. In a sense, our method is similar to theirs but from an RL perspective: we aim to find test environments that are on some boundaries. However, contrary to them, we do not have an actual test set to work on and have to generate the test environments. Moreover, we do not assume that the test environments on this boundary are more likely to reveal the mutation. Heuristically, we just assume it's a potential point of interest, better than any random point if we are constrained by the number of test environments to be generated and tested. Finally, Tambon et al \cite{Tambon22} study more in-depth the effect that the choice of Deep Learning model instances has over the result of Mutation Testing in Supervised Learning using DeepCrime's approach. They showed that this choice can affect the outcome of the Mutation Testing and that using the Bayesian approach can mitigate this issue. In our approach, we do not account for this effect and just focused on existing killing definitions and generation of test environments, since the main goal was to analyze FOM and subsequent HOM in RL. Nonetheless, it would be interesting to verify that a similar behavior also exists in RL.

\section{Conclusion}\label{sec:concl}
In this paper, we have presented \tool{}, a framework for Mutation Testing in RL. We have defined three distinct categories of FOM based on real faults that can happen during the design and training of deep RL agents. We compared different mutation killing definition choices based on the previous framework applying Mutation Testing to Deep Learning. We also evaluated HOM which are a combination of FOM and can prove to be more complex to kill which makes for interesting corner faults to investigate. The FOM used in the HOM generation were selected based on their relevance to generated parameterized test environments. We have tested our approach on a set of state-of-the-art RL algorithms (DQN, A2C, and PPO) over two benchmark environments (LunarLander and CartPole). The results of our study show that the mutation killing definition choice is important when it comes to killing mutations (\textit{ratio} vs. \textit{distribution} of rewards). We demonstrate that by testing the agents in modified environments, we can detect non-trivial FOM which are not detected by testing the agents in the environments they were trained in. Finally, we show that HOM generated from non-trivial FOM possess in the majority the interesting property of being subsuming, meaning that they are harder to kill than their constituent FOM and so more interesting from a testing perspective.

Overall this study showed that, while Mutation Testing applied to RL raises numerous challenges to consider from the mutation killing definition design to the generation of relevant test environments and HOM, it is an interesting venue to improve testing of RL-based software systems. As such, we believe that further research on this topic should focus on those issues to enhance Mutation Testing applied to RL.

\section*{Acknowledgment} 
This work was supported by: Fonds de Recherche du Québec (FRQ), the Canadian Institute for Advanced Research (CIFAR) as well as the DEEL project CRDPJ 537462-18 funded by the National Science and Engineering Research Council of Canada (NSERC) and the Consortium for Research and Innovation in Aerospace in Québec (CRIAQ), together with its industrial partners Thales Canada inc, Bell Textron Canada Limited, CAE inc and Bombardier inc.


\balance
\bibliography{refs}

\begin{thebibliography}{10}

\bibitem{Jia11}
Y.~Jia and M.~Harman, ``An analysis and survey of the development of mutation
  testing,'' {\em IEEE Transactions on Software Engineering}, vol.~37, no.~5,
  pp.~649--678, 2011.

\bibitem{Papadakis19}
M.~Papadakis, M.~Kintis, J.~Zhang, Y.~Jia, Y.~Le~Traon, and M.~Harman,
  ``Mutation testing advances: an analysis and survey,'' in {\em Advances in
  Computers}, vol.~112, pp.~275--378, Elsevier, 2019.

\bibitem{Offutt92}
A.~J. Offutt, ``Investigations of the software testing coupling effect,'' {\em
  ACM Trans. Softw. Eng. Methodol.}, vol.~1, p.~5–20, jan 1992.

\bibitem{Jia09}
Y.~Jia and M.~Harman, ``Higher order mutation testing,'' {\em Information and
  Software Technology}, vol.~51, no.~10, pp.~1379--1393, 2009.
\newblock Source Code Analysis and Manipulation, SCAM 2008.

\bibitem{Zhang22}
J.~M. Zhang, M.~Harman, L.~Ma, and Y.~Liu, ``Machine learning testing: Survey,
  landscapes and horizons,'' {\em IEEE Transactions on Software Engineering},
  vol.~48, no.~1, pp.~1--36, 2022.

\bibitem{Ma18}
L.~Ma, F.~Zhang, J.~Sun, M.~Xue, B.~Li, F.~Juefei-Xu, C.~Xie, L.~Li, Y.~Liu,
  J.~Zhao, {\em et~al.}, ``Deepmutation: Mutation testing of deep learning
  systems,'' in {\em 2018 IEEE 29th International Symposium on Software
  Reliability Engineering (ISSRE)}, pp.~100--111, IEEE, 2018.

\bibitem{Hu19}
Q.~Hu, L.~Ma, X.~Xie, B.~Yu, Y.~Liu, and J.~Zhao, ``Deepmutation++: A mutation
  testing framework for deep learning systems,'' in {\em 2019 34th IEEE/ACM
  International Conference on Automated Software Engineering (ASE)},
  pp.~1158--1161, IEEE, 2019.

\bibitem{humbatova2021deepcrime}
N.~Humbatova, G.~Jahangirova, and P.~Tonella, ``Deepcrime: mutation testing of
  deep learning systems based on real faults,'' in {\em Proceedings of the 30th
  ACM SIGSOFT International Symposium on Software Testing and Analysis},
  pp.~67--78, 2021.

\bibitem{Luong19}
N.~C. Luong, D.~T. Hoang, S.~Gong, D.~Niyato, P.~Wang, Y.-C. Liang, and D.~I.
  Kim, ``Applications of deep reinforcement learning in communications and
  networking: A survey,'' {\em IEEE Communications Surveys \& Tutorials},
  vol.~21, no.~4, pp.~3133--3174, 2019.

\bibitem{Polydoros17}
A.~S. Polydoros and L.~Nalpantidis, ``Survey of model-based reinforcement
  learning: Applications on robotics,'' {\em Journal of Intelligent \& Robotic
  Systems}, vol.~86, no.~2, pp.~153--173, 2017.

\bibitem{LavetDRL2018}
V.~François-Lavet, P.~Henderson, R.~Islam, M.~G. Bellemare, and J.~Pineau,
  ``{An Introduction to Deep Reinforcement Learning},'' {\em Foundations and
  Trends® in Machine Learning}, vol.~11, no.~3-4, pp.~219--354, 2018.

\bibitem{Lu2022-RLmutation}
Y.~Lu, W.~Sun, and M.~Sun, ``Towards mutation testing of reinforcement learning
  systems,'' {\em Journal of Systems Architecture}, vol.~131, p.~102701, 2022.

\bibitem{sutton2018reinforcement}
R.~S. Sutton and A.~G. Barto, {\em Reinforcement learning: An introduction}.
\newblock MIT press, 2018.

\bibitem{Goodfellow-et-al-2016}
I.~Goodfellow, Y.~Bengio, and A.~Courville, {\em Deep Learning}.
\newblock MIT Press, 2016.
\newblock \url{http://www.deeplearningbook.org}.

\bibitem{mnih2015human}
V.~Mnih, K.~Kavukcuoglu, D.~Silver, A.~A. Rusu, J.~Veness, M.~G. Bellemare,
  A.~Graves, M.~Riedmiller, A.~K. Fidjeland, G.~Ostrovski, {\em et~al.},
  ``Human-level control through deep reinforcement learning,'' {\em nature},
  vol.~518, no.~7540, pp.~529--533, 2015.

\bibitem{gandhi2017learning}
D.~Gandhi, L.~Pinto, and A.~Gupta, ``Learning to fly by crashing,'' in {\em
  2017 IEEE/RSJ International Conference on Intelligent Robots and Systems
  (IROS)}, pp.~3948--3955, IEEE, 2017.

\bibitem{moravvcik2017deepstack}
M.~Morav{\v{c}}{\'\i}k, M.~Schmid, N.~Burch, V.~Lis{\`y}, D.~Morrill, N.~Bard,
  T.~Davis, K.~Waugh, M.~Johanson, and M.~Bowling, ``Deepstack: Expert-level
  artificial intelligence in heads-up no-limit poker,'' {\em Science},
  vol.~356, no.~6337, pp.~508--513, 2017.

\bibitem{Jahangirova20}
G.~Jahangirova and P.~Tonella, ``An empirical evaluation of mutation operators
  for deep learning systems,'' in {\em 2020 IEEE 13th International Conference
  on Software Testing, Validation and Verification (ICST)}, pp.~74--84, 2020.

\bibitem{Nelder72}
J.~A. Nelder and R.~W.~M. Wedderburn, ``Generalized linear models,'' {\em
  Journal of the Royal Statistical Society. Series A (General)}, vol.~135,
  no.~3, pp.~370--384, 1972.

\bibitem{Kelley12}
K.~Kelley and K.~J. Preacher, ``On effect size,'' {\em Psychological Method},
  vol.~17, no.~2, pp.~137--152, 2012.

\bibitem{Henderson18}
P.~Henderson, R.~Islam, P.~Bachman, J.~Pineau, D.~Precup, and D.~Meger, ``Deep
  reinforcement learning that matters,'' in {\em Proceedings of the
  Thirty-Second AAAI Conference on Artificial Intelligence and Thirtieth
  Innovative Applications of Artificial Intelligence Conference and Eighth AAAI
  Symposium on Educational Advances in Artificial Intelligence},
  AAAI'18/IAAI'18/EAAI'18, AAAI Press, 2018.

\bibitem{Agarwal21}
R.~Agarwal, M.~Schwarzer, P.~S. Castro, A.~C. Courville, and M.~G. Bellemare,
  ``Deep reinforcement learning at the edge of the statistical precipice,'' in
  {\em NeurIPS}, 2021.

\bibitem{nikanjam2022faults}
A.~Nikanjam, M.~M. Morovati, F.~Khomh, and H.~Ben~Braiek, ``Faults in deep
  reinforcement learning programs: a taxonomy and a detection approach,'' {\em
  Automated Software Engineering}, vol.~29, no.~1, pp.~1--32, 2022.

\bibitem{humbatova2020taxonomy}
N.~Humbatova, G.~Jahangirova, G.~Bavota, V.~Riccio, A.~Stocco, and P.~Tonella,
  ``Taxonomy of real faults in deep learning systems,'' in {\em Proceedings of
  the ACM/IEEE 42nd International Conference on Software Engineering},
  pp.~1110--1121, 2020.

\bibitem{rep-package}
``{Replication Package}.'' \url{https://github.com/FlowSs/RLMutation}, 2022.

\bibitem{Everitt17}
T.~Everitt, V.~Krakovna, L.~Orseau, M.~Hutter, and S.~Legg, ``Reinforcement
  learning with a corrupted reward channel,'' 2017.

\bibitem{Romoff18}
J.~Romoff, P.~Henderson, A.~Piche, V.~Francois-Lavet, and J.~Pineau, ``Reward
  estimation for variance reduction in deep reinforcement learning,'' in {\em
  Proceedings of The 2nd Conference on Robot Learning} (A.~Billard, A.~Dragan,
  J.~Peters, and J.~Morimoto, eds.), vol.~87 of {\em Proceedings of Machine
  Learning Research}, pp.~674--699, PMLR, 29--31 Oct 2018.

\bibitem{Biagiola22}
M.~Biagiola and P.~Tonella, ``Testing the plasticity of reinforcement
  learning-based systems,'' {\em ACM Trans. Softw. Eng. Methodol.}, vol.~31,
  jul 2022.

\bibitem{stable-baselines3}
A.~Raffin, A.~Hill, A.~Gleave, A.~Kanervisto, M.~Ernestus, and N.~Dormann,
  ``Stable-baselines3: Reliable reinforcement learning implementations,'' {\em
  Journal of Machine Learning Research}, vol.~22, no.~268, pp.~1--8, 2021.

\bibitem{LunarLander}
``{Lunar Lander}.''
  \url{https://www.gymlibrary.dev/environments/box2d/lunar_lander/}, 2022.

\bibitem{cramer46}
H.~Cramer, {\em Mathematical Methods of Statistics}.
\newblock Princeton University Press, 1946.

\bibitem{gur2021environment}
I.~Gur, N.~Jaques, Y.~Miao, J.~Choi, M.~Tiwari, H.~Lee, and A.~Faust,
  ``Environment generation for zero-shot compositional reinforcement
  learning,'' {\em Advances in Neural Information Processing Systems}, vol.~34,
  pp.~4157--4169, 2021.

\bibitem{cobbe2020leveraging}
K.~Cobbe, C.~Hesse, J.~Hilton, and J.~Schulman, ``Leveraging procedural
  generation to benchmark reinforcement learning,'' in {\em International
  conference on machine learning}, pp.~2048--2056, PMLR, 2020.

\bibitem{Humeniuk22}
D.~Humeniuk, F.~Khomh, and G.~Antoniol, ``A search-based framework for
  automatic generation of testing environments for cyber–physical systems,''
  {\em Information and Software Technology}, vol.~149, p.~106936, 2022.

\bibitem{Shen21}
W.~Shen, Y.~Li, Y.~Han, L.~Chen, D.~Wu, Y.~Zhou, and B.~Xu, ``Boundary sampling
  to boost mutation testing for deep learning models,'' {\em Information and
  Software Technology}, vol.~130, p.~106413, 2021.

\bibitem{Tambon22}
F.~Tambon, F.~Khomh, and G.~Antoniol, ``A probabilistic framework for mutation
  testing in deep neural networks,'' {\em Information and Software Technology},
  vol.~155, p.~107129, 2023.

\end{thebibliography}
\bibliographystyle{ieeetr}

\end{document}